\documentclass[10pt,twocolumn,letterpaper]{article}

\usepackage[pagenumbers]{iccv} 
\usepackage{colortbl}

\usepackage[normalem]{ulem}
\usepackage{algorithm}
\usepackage{algpseudocode}

\definecolor{iccvblue}{rgb}{0.21,0.49,0.74}

\definecolor{deepskyblue}{rgb}{0.0, 0.75, 1.0}

\usepackage{multirow}
\usepackage{multicol}
\usepackage{amsthm}
\newtheorem{proposition}{Proposition}  
\usepackage{pifont}
\usepackage[pagebackref,breaklinks,colorlinks,allcolors=iccvblue]{hyperref}

\def\paperID{11811} 
\def\confName{ICCV}
\def\confYear{2025}

\title{TLB-VFI: Temporal-Aware Latent Brownian Bridge Diffusion for Video Frame Interpolation}

\author{Zonglin Lyu  \quad  Chen Chen\\
Center for Research in Computer Vision, University of Central Florida\\
{\tt\small zonglin.lyu@ucf.edu \quad chen.chen@crcv.ucf.edu}
}

\begin{document}

\twocolumn[{%
    \maketitle
    \renewcommand\tabcolsep{1pt}
    \centering

    \includegraphics[width=\linewidth]{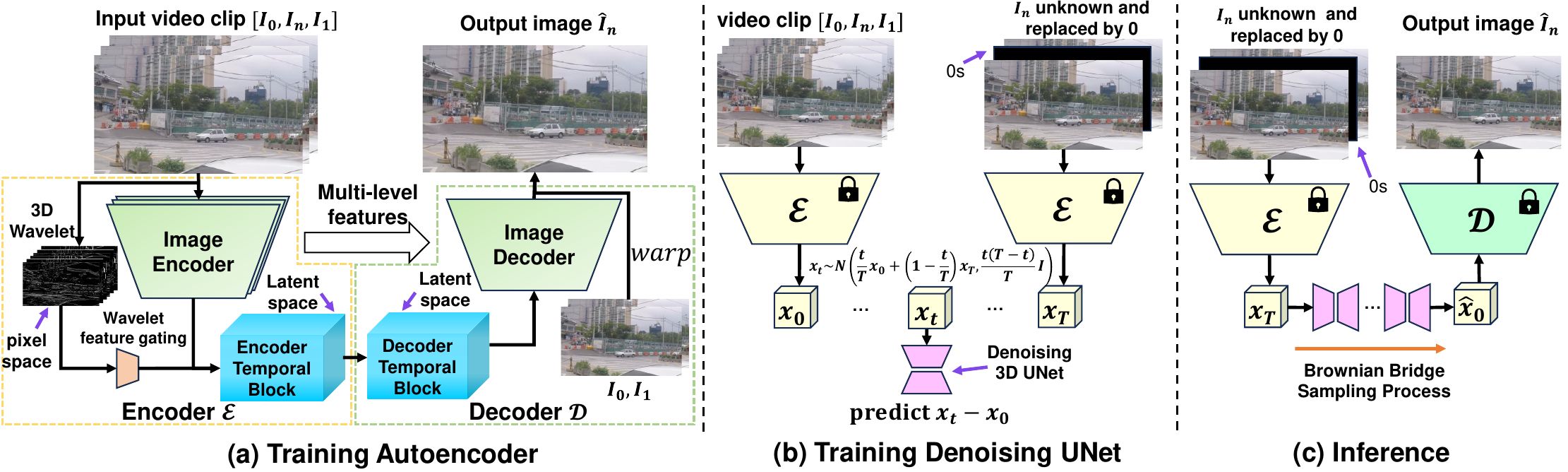} 
    \vspace{-6mm}
    \captionsetup{type = figure,font = footnotesize}
    \caption{
    \textbf{Overview of the proposed method.} \textbf{(a) Training autoencoder.} The autoencoder is trained with video clip $V = [I_0,I_n,I_1]$ and aims to reconstruct $I_n$. It contains an image encoder (shared for all frames) and an image decoder, where multi-level encoder features from $I_0,I_1$ are passed to the decoder. Temporal blocks extract temporal information in the latent space and aggregate video features into a single image feature for the image decoder, and 3D Wavelet extracts temporal information in the pixel space. \textbf{(b) Training Denoising UNet.} The video clip $V$ is encoded to $\textbf{x}_0$ by Encoder $\mathcal{E}$ (spatial + temporal). Since $I_n$ is unknown, we replace it by 0 and obtain another video clip $\tilde{V} = [I_0,0,I_1]$, which is encoded to $\mathbf{x}_T$. With the Brownian Bridge Diffusion Process, $\mathbf{x}_t$ is computed and sent to denoising UNet to predict $\mathbf{x}_t - \mathbf{x}_0$. \textbf{(c) Inference.} During inference, we encode $\tilde{V}$ to $\mathbf{x}_T$ and sample with the Brownian Bridge Sampling Process to get $\hat{\mathbf{x}}_0$, which is decoded to the output frame $\hat{I}_n$. \vspace{3mm}}
    \label{fig:teaser}

}]

\begin{abstract}
Video Frame Interpolation (VFI) aims to predict the intermediate frame $I_n$ (we use n to denote time in videos to avoid notation overload with the timestep $t$ in diffusion models) based on two consecutive neighboring frames $I_0$ and $I_1$. Recent approaches apply diffusion models (both image-based and video-based) in this task and achieve strong performance. However, image-based diffusion models are unable to extract temporal information and are relatively inefficient compared to non-diffusion methods. Video-based diffusion models can extract temporal information, but they are too large in terms of training scale, model size, and inference time. To mitigate the above issues, we propose Temporal-Aware Latent Brownian Bridge Diffusion for Video Frame Interpolation (TLB-VFI), an efficient video-based diffusion model. By extracting rich temporal information from video inputs through our proposed 3D-wavelet gating and temporal-aware autoencoder, our method achieves \textbf{20}\% improvement in FID on the most challenging datasets over recent SOTA of image-based diffusion models. Meanwhile, due to the existence of rich temporal information, our method achieves strong performance while having \textbf{3}$\times$ fewer parameters. Such a parameter reduction results in \textbf{2.3}$\times$ speed up. By incorporating optical flow guidance, our method requires \textbf{9000}$\times$ less training data and achieves over \textbf{20}$\times$ fewer parameters than video-based diffusion models. Codes and results are available at our \textcolor{iccvblue}{\href{https://zonglinl.github.io/tlbvfi_page}{Project Page}}.
\vspace{-3mm}
\end{abstract}
    
\section{Introduction}
\label{sec:intro}

Video Frame Interpolation (VFI) is a crucial task in computer vision that aims to predict the intermediate frame $I_n$ between two known neighboring frames (past frame $I_0$ and future frame $I_1$). Good quality interpolation results benefit a wide range of applications, including novel-view synthesis~\cite{flynn2016deepstereo} and video compression~\cite{wu2018video}.

Recent VFI methods generally fall into two categories: diffusion-based and traditional (a.k.a non-diffusion-based) methods. Traditional methods use either kernel-based methods to generate convolution kernels that predict intermediate frames from pixels of neighboring images~\cite{cheng2020video,lee2020adacof,niklaus2017videosep,shi2021video,niklaus2017video} or flow-based methods that estimate optical flows~\cite{lu2022video,plack2023frame,argaw2022long,huang2022real,dutta2022non,choi2021high,licvpr23amt,park2023BiFormer,zhang2023extracting,zhang2024vfimamba,wu2024perception}. Advances in deep learning make researchers prefer flow-based methods over kernel-based ones because the motion estimation in flow-based methods benefit from deep architecture, whereas kernel-based methods do not model motion explicitly. Diffusion-based methods can be split into image-based diffusion models~\cite{danier2024ldmvfi,lyu2024frame} and video-based diffusion models~\cite{voleti2022mcvd,jain2024video,shen2024dreammover,yang2024vibidsampler}, both achieving strong performances. However, they encounter several limitations:

\begin{itemize}
    \item Traditional methods and image-based diffusion methods (generally less efficient than traditional methods)~\cite{lyu2024frame,danier2024ldmvfi}, which incorporate kernel- or flow-based methods to explicitly guide frame interpolation, only extract spatial information in $I_0,I_1$. Therefore, \uline{they lack explicit temporal information extraction to enhance temporal consistency}.
    
    \item Video-based diffusion models~\cite{voleti2022mcvd,jain2024video,shen2024dreammover,yang2024vibidsampler} directly generate pixels. Though they successfully extract temporal information by taking video clips as inputs, \uline{the training and inference costs of these methods are extremely high}. They require more than ten million videos to train for a strong performance due to the lack of explicit pixel-level information from $I_0,I_1$ guided by optical flow, and the inference time is also extremely long.
    
\end{itemize}

Our goal is \uline{to extract temporal information while keeping training scale, model size, and inference time reasonable}. To achieve so, we notice that a video-based diffusion model is necessary. Video-based diffusion models can gradually build up temporal information in their sampling process with 3D UNet, and the autoencoder can further extract temporal information. This is not feasible in traditional methods as $I_n$ is not available as input. Therefore, we propose Temporal-Aware Latent Brownian Bridge Diffusion for Video Frame Interpolation (TLB-VFI), shown in Fig.~\ref{fig:teaser}. Starting with the selection of diffusion models, Consec. BB~\cite{lyu2024frame} points out that Brownian Bridge Diffusion Model (BBDM)~\cite{li2023bbdm} better fits the VFI task than traditional Diffusion Models~\cite{ho2020denoising} since the sampling variance of BBDM is lower. This is because VFI expects deterministic interpolation results rather than diverse ones. However, in the setup of Consec. BB~\cite{lyu2024frame}, Brownian Bridge is applied between adjacent frames, \uline{but the latent features are almost identical for adjacent frames}. Therefore, the Brownian Bridge becomes approximately an identity mapping and loses its functionality. To mitigate this problem, given input video clip $V = [I_0,I_n,I_1]$, we construct $\tilde{V} = [I_0,0,I_1]$ and apply Brownian Bridge Diffusion between their latent features, shown in Fig.~\ref{fig:teaser} (b). The latent features of $\tilde{V}$ largely differ from that of $V$, resolving the issue of identity mapping. Details and justifications are in Sec.~\ref{sec:ablatoin} and Fig.~\ref{fig:mse}.

Next, it is important to construct an autoencoder so that diffusion models can efficiently run in latent space. \textbf{To overcome the limitation on lack of temporal information}, a simple approach is to use 3D Convolution and Spatiotemporal attention. However, since $I_n$ is replaced with 0 during inference (see Fig.~\ref{fig:teaser} (c)), the multi-level encoder features, which benefits the decoding stage~\cite{danier2024ldmvfi}, in Fig.~\ref{fig:teaser} (a) will contain incomplete information due to zero replacement, which harms the performance (see Sec.~\ref{sec:ablatoin}). We mitigate this issue by \uline{splitting the temporal information extraction into pixel space and latent space} as shown in Fig.~\ref{fig:teaser} (a). We include image-level encoder and decoder for spatial features only and add temporal feature extraction between the image-level encoder and decoder to extract temporal information in the latent space. This enables us to utilize the multi-level encoder features of $I_0,I_1$ during the decoding phase as \textit{they are not impacted by zero replacement}. To extract temporal information in the pixel space, we propose our 3D-wavelet feature gating mechanism to extract high-frequency temporal information, since high-frequency information represents the changes along temporal dimension. \textbf{To address the limitation on efficiency}, instead of directly predicting pixel values, we utilize optical flow estimation to warp neighboring frames and refine the output. Under the guidance of optical flow at the pixel level in $I_0$ and $I_1$, our method achieves better efficiency than directly generating pixels from scratch. Importantly, due to rich temporal features extracted, our method achieves the state-of-the-art performance while requiring less parameters and inference time than image-based diffusion models~\cite{lyu2024frame,danier2024ldmvfi}. Our contributions are summarized as:

\begin{itemize}
    \item Our method requires \textbf{9000}$\times$ fewer training data, has over \textbf{20}$\times$ fewer the number of parameters, and is over \textbf{10}$\times$ faster than video-based diffusion models~\cite{jain2024video,yang2024vibidsampler,shen2024dreammover}. Comparing to Image-based diffusion methods, our method has \textbf{3}$\times$ fewer parameters and over \textbf{2}$\times$ faster.
    
    \item We introduce our temporal design of autoencoder to extract temporal information in the latent space and 3D wavelet feature gating to extract temporal information in the pixel space, serving as complement with each other. 
    \item We propose a theoretical constraint on \textit{when Brownian Bridge Diffusion is effective} in Sec.~\ref{sec:temporal}. Our temporal design of autoencoder adheres to our proposed constraint, validating the effectiveness of the Brownian Bridge.
    
    \item Through extensive experiments, our method achieves state-of-the-art performance in various datasets. Specifically, our method achieves around \textbf{20\%} improvement in FID in the most challenging datasets: SNU-FILM extreme subset~\cite{choi2020channel} and Xiph-4K~\cite{Niklaus_CVPR_2020} over recent SOTAs.
\end{itemize}

\section{Related Work}
\label{sec:related}

\subsection{Traditional Methods in VFI}

Video Frame Interpolation (VFI) is a task to predict the intermediate frame $I_n$ given its neighboring frames $I_0$ and $I_1$. Traditional (non-diffusion) VFI methods fall into two categories: flow-based methods~\cite{plack2023frame,lu2022video,jin2023unified,huang2022real,argaw2022long,choi2021high,dutta2022non,licvpr23amt,park2023BiFormer,zhang2023extracting} and kernel-based methods~\cite{cheng2020video,lee2020adacof,niklaus2017video,niklaus2017videosep,shi2021video}. Flow-based methods utilize optical flow estimations via deep neural networks. Some estimate optical flows from the $I_n$ to $I_0$ and $I_1$ and apply backward warping~\cite{lu2022video,plack2023frame,huang2022real,dutta2022non,choi2021high, licvpr23amt,park2023BiFormer,argaw2022long,zhang2023extracting}. Others estimate bidirectional flows from $I_0,I_1$ to each other, and apply forward splatting~\cite{jin2023unified}. In addition to the basic framework, various architectures like transformers~\cite{vaswani2017attention} are introduced together with techniques such as multi-resolution recurrence refinement~\cite{jin2023unified}, 4D-correlations extraction~\cite{licvpr23amt}, and asymmetric flow blending~\cite{wu2024perception} to enhance interpolation quality. Kernel-based methods, initially proposed by~\cite{niklaus2017video}, try to estimate local convolution kernels that are applied to pixels of neighboring frames to generate pixels in the intermediate frame. Several improvements, such as adaptive and deformable convolutions~\cite{cheng2020video,lee2020adacof}, are proposed to improve performance under large motion change. 

\subsection{Diffusion Models in VFI}

Denoising Diffusion Probabilistic Models (DDPM)~\cite{ho2020denoising} are proposed to generate realistic and diverse images. In DDPM, an image is transformed into standard Gaussian noise with a predefined diffusion process, and the noise is transformed back to an image with a sampling (denoising) process. The sampling process contains T steps of iteration, where T is set to 1000 experimentally. The following works~\cite{song2021denoising,lu2022dpm,lu2022dpm++} improve the efficiency of sampling by reducing the number of steps for iterative sampling while keeping high-quality generation. Latent Diffusion Models~\cite{rombach2022high} introduce autoencoders with vector-quantized or KL-divergence regularization to compress images into latent space, where diffusion models operate more efficiently. Beyond image generation, BBDM~\cite{li2023bbdm} proposes Brownian Bridge Diffusion for image-to-image translation, and video diffusion models~\cite{ho2022video,blattmann2023stable} are developed to generate high-quality videos.

LDMVFI~\cite{danier2024ldmvfi} incorporates the idea of Latent Diffusion Models (LDMs)~\cite{rombach2022high} and kernel-based methods, proposing an autoencoder that utilizes kernel-based methods to reconstruct $I_n$. Recent work~\cite{lyu2024frame} claims that VFI prefers deterministic prediction of the intermediate frames and introduces the Consecutive Brownian Bridge, which has a small sampling variance and achieves state-of-the-art performance. Other diffusion-based works~\cite{jain2024video,voleti2022mcvd,shen2024dreammover,yang2024vibidsampler} do not employ kernel- or flow-based methods but instead generate raw pixels. This approach allows these diffusion methods to extract explicit temporal information by taking videos as inputs to autoencoder and build up temporal information during the sampling process from a random noise. However, these methods require large-scale training as they lack pixel guidance from neighboring frames $I_0,I_1$: most works~\cite{jain2024video,shen2024dreammover,yang2024vibidsampler} require over ten million videos to train, while the common datasets for VFI only contain 51K triplets~\cite{xue2019video}. MCVD~\cite{voleti2022mcvd}, a diffusion approach directly generating pixels, is trained with a small-scale dataset, resulting in unsatisfactory performance in VFI~\cite{danier2024ldmvfi}. Our method addresses these challenges by extracting temporal information from video inputs and taking advantage of optical flow estimation.
\begin{figure*}[t]
\centering
\includegraphics[width=\textwidth]{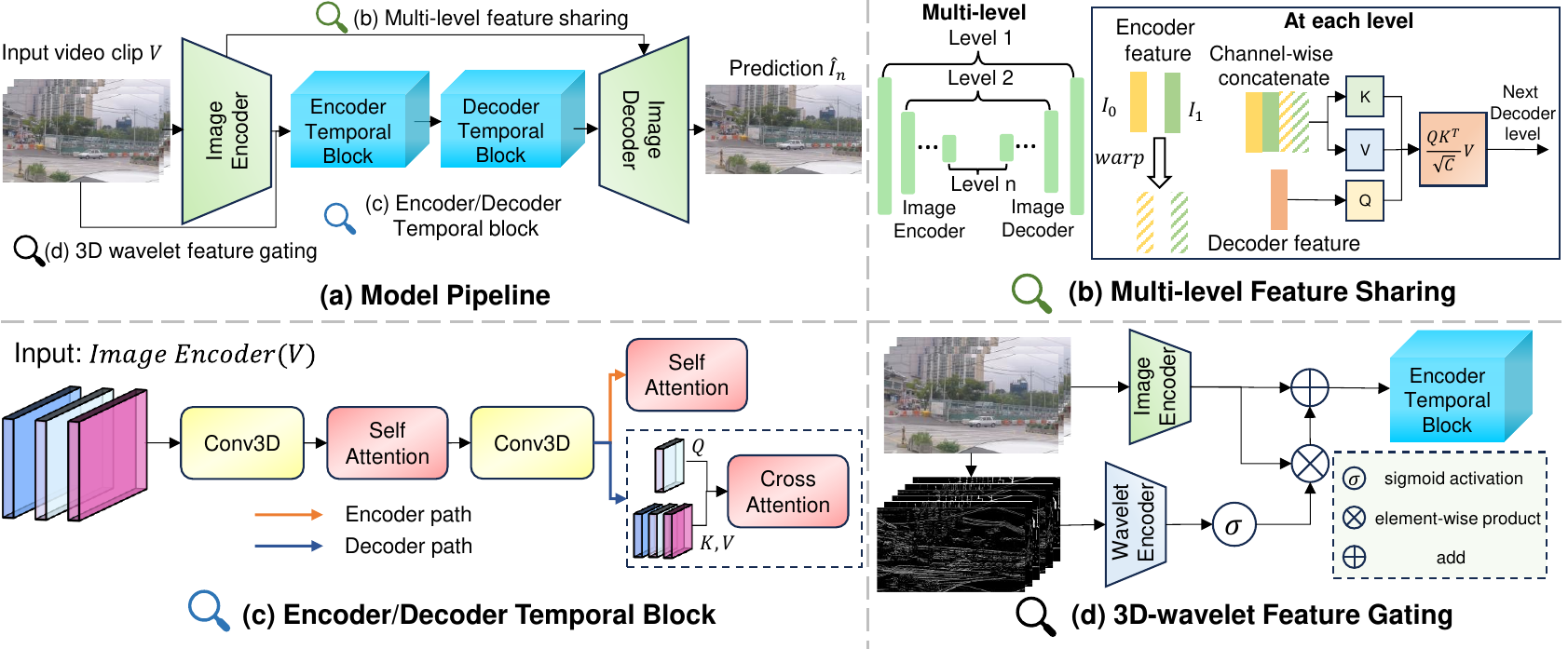}
\captionsetup{font=footnotesize,aboveskip=1pt}
\caption{\textbf{(a) Model Pipeline.} The Image Encoder is shared across all frames, and temporal blocks extract temporal information in the latent space. \textbf{(b) Multi-level Feature Sharing.} The Image Encoder and Decoder consist of several levels of resolution due to downsampling/upsampling latent features. At the $i^{th}$ level of the encoder and the decoder, features from $I_0$ and $I_1$ in the encoder are warped and concatenated with the original copy (when the downsampling rate is larger than 8, warped features are excluded). The concatenated features are used as keys and values in cross attention~\cite{vaswani2017attention} where the decoder feature at the same level is the query. \textbf{(c) Encoder/Decoder Temporal Block.} Each temporal block consists of two sets of 3D convolution + attention. In the decoder, the second attention is cross-attention between the intermediate frame (query) and all frames (key and value) to aggregate the video feature into one feature map. \textbf{(d) 3D-wavelet Feature Gating.} Wavelet information is extracted from the input video clip and encoded by CNNs. A sigmoid activation is applied, and the result is element-wise multiplied by the output of the Image Encoder with a skip connection.\vspace{-3mm}}
\label{fig:arch}
\end{figure*}

\section{Methodology}

\subsection{Preliminary}

\textbf{Diffusion Models.} DDPM~\cite{ho2020denoising} defines a diffusion process that converts images into standard Gaussian noise:

\begin{equation}
    q(\mathbf{x}_t|x_0) = \mathcal{N}(x_t;\sqrt{\alpha}_t\mathbf{x}_0,(1-\alpha_t)\mathbf{I}),
    \label{eq: ddpm marginal}
\end{equation}
\begin{equation*}
    \text{where }\alpha_t = \prod_{s=1}^t(1-\beta_s).
\end{equation*}
$\{\beta_t\}$ are small pre-defined constants. The sampling process (i.e. how to denoise $\mathbf{x}_{t-1}$ from $\mathbf{x}_t$) is given by~\cite{ho2020denoising}:
\begin{equation}
    \label{eq:ddpm sample}
    p_\theta(\mathbf{x}_{t-1}|\mathbf{x}_t) = \mathcal{N}(x_{t-1};\tilde{\mathbf{\mu}}_t,\tilde{\beta}_t),
\end{equation}
\begin{equation}
    \text{where }\tilde{\mathbf{\mu}}_t = \frac{1}{1-\beta_t}\left(\mathbf{x}_t - \frac{\beta_t}{\sqrt{1-\alpha_t}}\epsilon\right),
    \label{eq:ddpm sample mean}
\end{equation}
\begin{equation}
     \tilde{\beta}_t = \frac{1-\alpha_{t-1}}{1-\alpha_t}\beta_t, \text{ and } \epsilon\sim\mathcal{N}(0,\mathbf{I}).
\end{equation}

Based on Eq.~\eqref{eq:ddpm sample mean}, the only unknown term is $\epsilon$, which is estimated with a deep neural network $\epsilon_\theta(\mathbf{x}_t,t)$.

\textbf{Brownian Bridge Diffusion Models.} Similar to DDPM, Brownian Bridge Diffusion Models (BBDM)~\cite{li2023bbdm} replace its diffusion and sampling process with Brownian Bridge Diffusion~\cite{ross1995stochastic} for image-to-image translation tasks such as image inpainting and image colorization. Consecutive Brownian Bridge~\cite{lyu2024frame} introduces a cleaner formulation of BBDM and suggests that it fits the VFI task since the sampling process has low variance, which VFI prefers. The diffusion process is:
\begin{equation}
    \label {eq:BBDM forward}
    \small
    q(\mathbf{x}_t|\mathbf{x}_0,\mathbf{x}_T) = \mathcal{N}\left(\frac{t}{T}\mathbf{x}_0 + (1-\frac{t}{T})\mathbf{x}_T,\frac{t(T-t)}{T}\mathbf{I}\right).
\end{equation}

$\mathbf{x}_0$ and $\mathbf{x}_T$ are the latent features of two images to be translated. $T$ controls the maximum variance of the Brownian Bridge. The sampling process is given as~\cite{lyu2024frame}:
\begin{equation}
\label{BBDM_sample}
p_\theta(\mathbf{x}_{s}|\mathbf{x}_t,\mathbf{x}_T) = \mathcal{N}\left(\mathbf{x}_t - \frac{\Delta_t
}{t}(\mathbf{x}_t - \mathbf{x}_0),\frac{s\Delta_t}{t}\mathbf{I}\right).
\end{equation}

Here $s$ is an arbitrary time before $t$, and $\Delta_t = t-s$. Analogous to DDPM~\cite{ho2020denoising}, the only unknown term is $\mathbf{x}_t - \mathbf{x}_0$, which can be estimated with a deep neural network.

\subsection{Problem Definition}

Video Frame Interpolation aims to predict $I_n \in \mathbb{R}^{3\times H \times W}$, which is the intermediate frames between $I_0$ and $I_1$. Following the common strategy of recent VFI works~\cite{lyu2024frame,wu2024perception}, the goal is to predict $\hat{I}_n$ that predicts $I_n$:
\begin{equation}
    \hat{I}_n = M\odot warp(I_0) + (1-M)\odot warp(I_1) + \Delta
\end{equation}

$M$ is a mask with values between 0 and 1. $warp$ denotes warping based on estimated optical flows. $\Delta$ is the residual term with values between -1 and 1. Therefore, our model will predict the mask $M$ and residual $\Delta$, with flow estimation embedded in the model. The basics of optical flow and warping are included in the \textcolor{blue}{Supplementary Material Sec.~\ref{sec: flow}}.

\textbf{Method Pipeline.} Our method is divided into two main components: an autoencoder and a Brownian Bridge Diffusion Model (referred to as `diffusion model' later), shown in Fig.~\ref{fig:teaser} (a) and (b), respectively. The autoencoder contains an encoder $\mathcal{E}$ and a decoder $\mathcal{D}$, shown in \textcolor{yellow}{yellow} and \textcolor{green}{green} dashes in Fig.~\ref{fig:teaser} (a) respectively. The encoder $\mathcal{E}$ contains an image encoder and temporal blocks, and the decoder contains temporal blocks and an image decoder. The reason for this design is discussed in Sec.~\ref{sec:temporal}. The objective of the autoencoder is to predict $M,\Delta$ given input video clip $V = [I_0,I_n,I_1]$: $M,\Delta$ = $\mathcal{D}(\mathcal{E}(V))$. During inference, $I_n$ is replaced with a zero matrix, resulting in $\tilde{V}$ = $[I_0,0,I_1]$. The goal of the diffusion model (denoted as $BB$) is to estimate $\mathcal{E}(V)$ from $\mathcal{E}(\tilde{V})$. In summary, our pipeline to predict $I_n$, with $n=0.5$, involves the following steps:

\begin{enumerate}
    \item $x = BB(\mathcal{E}(\tilde{V})).$\vspace{1mm}
    \item $M,\Delta = \mathcal{D}(x).$\vspace{1mm}
    \item $\hat{I}_n = M\odot warp(I_0) + (1-M)\odot warp(I_1) + \Delta$
\end{enumerate}

\subsection{Temporal Aware Latent Brownian Bridge}

\textbf{Temporal Feature Extraction in Latent Space.} Our goal is to extract temporal information in the autoencoder, but it is important to provide multi-level encoder features of $I_0,I_1$ to guide the decoder~\cite{lyu2024frame,danier2024ldmvfi}. Details of how these multi-level features can provide guidance are shown in Fig.~\ref{fig:arch} (b). At $i^{th}$ level of the encoder and decoder, the features are downsampled to $2^{-i}\times$ of the image's height and width.  We denote encoder features of $I_0,I_1$ at $i^{th}$ level as $f_0,f_1$. Then $concat(f_0,f_1,warp(f_0),f_1,warp(f_1))$ is sent to the decoder at the $i^{th}$ level via cross-attention~\cite{vaswani2017attention} ($concat$ is channel-wise concatenation). To ensure multi-level encoder features of $I_0,I_1$ are not affected by changing $V$ to $\tilde{V}$ during inference, we use shared image encoders to encode each frame into low-resolution latent features and apply temporal feature extraction of those latent features. The temporal blocks for the encoder, shown in Fig.~\ref{fig:arch} (c), utilize 3D convolution and spatiotemporal attention. 

The decoder is mostly symmetric with the encoder, beginning with 3D convolution and spatiotemporal attention followed by an image decoder. To convert the video latent representation into the image latent representation, we include a spatiotemporal cross-attention aggregation in the decoder, illustrated in Fig.~\ref{fig:arch} (c). Given the video latent representation $F \in \mathbb{R}^{C\times T \times H \times W}$ where $C,T,H,W$ represents the channel, temporal, height, and weight. The spatiotemporal cross-attention aggregation is defined as:

\begin{equation*}
    V_{out} = softfmax\left(\frac{QK^T}{\sqrt{C}}\right)V
\end{equation*}

\vspace{-4mm}
\begin{equation*}
    Q = W_QF_Q, K = W_KF_{KV}, V = W_VF_{KV}
\end{equation*}
\vspace{-4mm}
\begin{equation*}
    F_Q = F[t].flatten(1), F_{KV} = F.flatten(1)
\end{equation*}

$F_Q$ and $F_{KV}$ are in PyTorch style notation. This approach allows us to remove the temporal dimension, retaining only the image-level feature. 

\textbf{Temporal Feature Extraction in Pixel Space with 3D Wavelet Transform.} 
\label{sec:temporal}
Since we use a shared image encoder to encode images, the encoder and decoder only extract temporal information in latent space. However, extracting temporal information at the pixel level is essential as the latent space is highly compressed. Therefore, we apply 3D wavelet transform~\cite{wavelet} to extract spatial and temporal frequency information with low-pass filter $[\frac{1}{\sqrt{2}} \frac{1}{\sqrt{2}}]$ and high-pass filter $[-\frac{1}{\sqrt{2}} \frac{1}{\sqrt{2}}]$. These filters are applied on height, width, and temporal dimensions, respectively. The reason to choose 3D wavelet transform is that it can apply different combinations of high pass and low pass filters across different dimensions, resulting in rich pixel-level information. Then, the frequency information is encoded with convolution layers. The 3D wavelet transform can be applied twice. With the first pass, we can extract temporal information between $I_0$ and $I_n$ and between $I_n$ and $I_1$, which is not achievable with input only containing $I_0$ and $I_1$. The second pass extracts the temporal information throughout all frames. Such pixel-level temporal information tells the model \textit{where the motion changes more drastically}, and with this intuition, we design our 3D wavelet feature gating, shown in Fig.~\ref{fig:arch} (d). If $f_w$ is the encoded feature of frequency information, and if $f_i$ is the latent features encoded by the shared image encoder, then the 3D wavelet gating is:
\begin{equation}
\label{eq:gating}
    f = \sigma(f_w)\odot f_i + f_i,
\end{equation}
where $\sigma$ is the sigmoid activation, and $\odot$ is the element-wise product. This implicitly guides the model to learn which parts of the video clip have more changes in the temporal dimension. The details about the implementation of the wavelet transform are included in the \textcolor{blue}{Supplementary Material Sec.~\ref{sec:wavelet}}.

\begin{table*}[t]
    \centering
    \captionsetup{font=footnotesize}
    \caption{Quantitative results in LPIPS$\downarrow$/FloLPIPS$\downarrow$/FID$\downarrow$, the lower the better, on evaluation datasets. The best performances are \textbf{\textcolor{blue}{blue-boldfaced}}, and the second-best performances are \underline{\textcolor{red}{red-underlined}}. Results from PerVFI are in \colorbox{gray!30}{gray background} and excluded from the ranking because their training scale is significantly larger. Results for baselines, except for PerVFI, are adopted from Consec.BB~\cite{lyu2024frame}. OOM indicates that the inference with one image exceeds the 24GB GPU memory of an Nvidia RTX A5000 GPU. Runtime measures seconds per frame to interpolate a $480\times720$ image with A5000 GPU. To ensure a fair comparison, we report the runtime of Consec. BB~\cite{lyu2024frame} and LDMVFI~\cite{danier2024ldmvfi} in 10 step sampling (our setup) and also report the runtime of their setup in \textcolor{orange}{orange} and parenthesis. Note that results from~\cite{jain2024video,shen2024dreammover,yang2024vibidsampler} are not included as their training scales are millions of videos. \vspace{-3mm}} 
    \label{tab:results}
        \resizebox{\textwidth}{!}{
        \begin{tabular}{lcccccccc}
        \toprule
       \multirow{2}{*}{Methods} &\multirow{2}{*}{Xiph-4K} & \multirow{2}{*}{Xiph-2K}& \multirow{2}{*}{DAVIS}& \multicolumn{4}{c}{SNU-FILM} &\multirow{2}{*}{Runtime} \\
        \cmidrule{5-8}
        & & & &easy & medium & hard&extreme\\
        & LPIPS/FloLPIPS/FID &LPIPS/FloLPIPS/FID&LPIPS/FloLPIPS/FID&LPIPS/FloLPIPS/FID&LPIPS/FloLPIPS/FID&LPIPS/FloLPIPS/FID&LPIPS/FloLPIPS/FID & Seconds\\
        \midrule
        
        MCVD'22~\cite{voleti2022mcvd} &OOM &OOM& 0.247/0.293/28.002 &  0.199/0.230/32.246 & 0.213/0.243/37.474 &0.250/0.292/51.529&0.320/0.385/83.156&52.55\\

        VFIformer~\cite{lu2022video} & OOM &OOM & 0.127/0.184/14.407 &0.018/0.029/5.918 &0.033/0.053/11.271 &0.061/0.100/22.775 &0.119/0.185/40.586 &4.34 \\
        
        IFRNet'22~\cite{Kong_2022_CVPR} &0.136/0.164/\textcolor{red}{\uline{23.647}}&0.068/0.093/\textcolor{red}{\uline{11.465}}&0.106/0.156/12.422 & 0.021/0.031/6.863&0.034/0.050/12.197 &0.059/0.093/23.254 &0.116/0.182/42.824 &0.10\\

        AMT'23~\cite{licvpr23amt} &0.199/0.230/29.183&0.089/0.126/13.100 & 0.109/0.145/13.018 &0.022/0.034/6.139 & 0.035/0.055/11.039 & 0.060/0.092/20.810&0.112/0.177/40.075 &0.11\\
        
        UPR-Net'23~\cite{jin2023unified} & 0.230/0.269/31.043 &0.103/0.144/12.909 & 0.134/0.172/15.002 & 0.018/0.029/5.669&0.034/0.052/10.983  &0.062/0.097/22.127 & 0.112/0.176/40.098 &0.70 \\
        
        EMA-VFI'23~\cite{zhang2023extracting} & 0.241/0.260/28.695 & 0.110/0.132/12.167& 0.132/0.166/15.186 & 0.019/0.038/5.882 & 0.033/0.053/11.051 & 0.060/0.091/20.679& 0.114/\textcolor{red}{\underline{0.170}}/39.051&0.72\\
        
        LDMVFI'24~\cite{danier2024ldmvfi} & OOM & OOM & 0.107/0.153/12.554 & 0.014/0.024/5.752 & 0.028/0.053/12.485 & 0.060/0.114/26.520& 0.123/0.204/47.042&2.48 (\textcolor{orange}{22.32})\\
        \rowcolor{gray!30}
        PerVFI'24~\cite{wu2024perception} &0.086/0.128/18.852 &0.038/0.069/10.078 & {0.081/0.122/8.217}&{0.014/0.022/5.917} &{0.024/0.040/10.395} &{0.046/0.077/18.887} &{0.090/0.151/32.372}&1.52\\
        
        Consec. BB'24~\cite{lyu2024frame} &\textcolor{red}{\underline{0.097}}/\textcolor{red}{\underline{0.135}}/24.424 &\textcolor{red}{\underline{0.042}}/\textcolor{red}{\underline{0.080}}/12.011& \textcolor{red}{\underline{0.092}}/\textcolor{red}{\underline{0.136}}/\textcolor{red}{\underline{9.220}}&\textcolor{red}{\underline{0.012}}/\textcolor{red}{\underline{0.019}}/\textcolor{red}{\underline{4.791}} &\textcolor{red}{\underline{0.022}}/\textcolor{red}{\underline{0.039}}/\textcolor{red}{\underline{9.039}} &\textcolor{red}{\underline{0.047}}/\textcolor{red}{\underline{0.091}}/\textcolor{red}{\underline{18.589}} &\textcolor{red}{\underline{0.104}}/0.184/\textcolor{red}{\underline{36.631}} & 1.62 (\textcolor{orange}{2.60})\\

        \midrule
        Ours &\textcolor{blue}{\textbf{0.077}}/\textcolor{blue}{\textbf{0.113}}/\textcolor{blue}{\textbf{19.114}} &\textcolor{blue}{\textbf{0.032}}/\textcolor{blue}{\textbf{0.067}}/\textcolor{blue}{\textbf{9.901}} & \textcolor{blue}{\textbf{0.086}}/\textcolor{blue}{\textbf{0.126}}/\textcolor{blue}{\textbf{8.299}}&\textcolor{blue}{\textbf{0.012}}/\textcolor{blue}{\textbf{0.018}}/\textcolor{blue}{\textbf{4.658}} &\textcolor{blue}{\textbf{0.021}}/\textcolor{blue}{\textbf{0.036}}/\textcolor{blue}{\textbf{8.518}} &\textcolor{blue}{\textbf{0.044}}/\textcolor{blue}{\textbf{0.085}}/\textcolor{blue}{\textbf{17.470}} &\textcolor{blue}{\textbf{0.095}}/\textcolor{blue}{\textbf{0.151}}/\textcolor{blue}{\textbf{29.868}}&0.69\\
        \bottomrule
    \end{tabular}
  }
  \vspace{-5mm}
\end{table*}

\textbf{Temporal Information Restoration with Brownian Bridge Diffusion.} During inference, $V$ (original video clip) is replaced with $\tilde{V}$ ($I_n$ replaced by zero matrix), leading to a distribution shift of encoded features $\mathcal{E}(\tilde{V})$ from $\mathcal{E}(V)$ due to loss of temporal information. We employ the Brownian Bridge Diffusion to align the distribution of $\mathcal{E}(v)$ and $\mathcal{E}(\tilde{V})$ to restore such information because Brownian Bridge diffusion demonstrates efficacy in VFI~\cite{lyu2024frame} by achieving low sampling variance. In addition, restoring the temporal information is conceptually similar to spatial information restoration in image inpainting in BBDM~\cite{li2023bbdm}, where the Brownian Bridge Diffusion is originally proposed. To effectively restore temporal information, temporal features need to be extracted. Therefore, we replace the convolution in the denoising U-Net with 3D convolution, and self-attentions are performed in spatial and temporal dimensions.

Notably, by Eq.~\eqref{BBDM_sample}, the denoising UNet $\epsilon_\theta$ aims to predict $\mathbf{x}_t - \mathbf{x}_0$. When $t = T$, the $\epsilon_\theta$ aims to predict $\mathbf{x}_T - \mathbf{x}_0$. If $\mathbf{x}_T = \mathbf{x}_0$, then by Eq.~\eqref{eq:BBDM forward}, $\mathbb{E}(\mathbf{x}_t) = 0$ at any time step. The training objective of $\epsilon_\theta$ becomes 0 on expectation based on Eq.~\eqref{BBDM_sample}. The variance is negligible because $\Delta_t = \frac{1}{1000}$ and $T = 2$ during training~\cite{lyu2024frame}. In this case, when $\epsilon_\theta$ learns to predict 0, we can easily compute that the sampling process defined by Eq.~\eqref{BBDM_sample} is an identity mapping with a small variance. Therefore, we need a significantly large distribution shift between $\mathbf{x}_0$ and $\mathbf{x}_T$ to prevent such a problem of identity mapping. Specifically, a big shift in the mean is required, which is described in the following proposition:

\begin{proposition}
\label{propostion}
    If the Brownian Bridge Diffusion is applied to translate between two distributions $\mathbf{x}_0$ and $\mathbf{x}_T$, there should be a large shift between $\mathbb{E}(\mathbf{x}_0)$ and $\mathbb{E}(\mathbf{x}_T)$. A sufficient constraint is to reject the Null Hypothesis $H_0:\mathbb{E}(\mathbf{x}_0 - \mathbf{x}_T) = 0$ at significance level $\alpha$.
\end{proposition}

The proof of this proposition is given in the \textcolor{blue}{Supplementary Material Sec.~\ref{sec:proof}}. We include experiments in Sec.~\ref{sec:ablatoin} to show that our method satisfies this constraint, while the Consec. BB~\cite{lyu2024frame} does not.
\section{Experiments}
\label{sec:experiment}

\subsection{Datasets, Evaluation Metrics, and Baselines}
\textbf{Datasets.} Following recent works~\cite{zhang2023extracting,lyu2024frame}, our autoencoder and diffusion models are trained on the Vimeo 90K triplets dataset~\cite{xue2019video}, which comprises 51,312 training triplets. We employ random flipping, cropping, rotation, and temporal order reversing as data augmentation. To evaluate our methods, we include Xiph~\cite{Niklaus_CVPR_2020}, DAVIS~\cite{Perazzi_CVPR_2016}, and SNU-FILM~\cite{choi2020channel}. SNU-FILM contains four subsets based on the magnitude of motion changes: easy, medium, hard, and extreme, and Xiph contains 4K and 2K resolution for evaluation. These datasets contain various resolutions (480p to 4K) and motion changes. Among these datasets, Xiph and SNU-FILM-extreme are the most challenging and important due to high resolution and extremely large motion changes respectively, and strong performance on these two datasets indicates strong real-world applicability.

\textbf{Evaluation Metric and Baselines.} Recent works~\cite{danier2024ldmvfi,lyu2024frame,wu2024perception} identify that pixel-based metrics such as PSNR and SSIM~\cite{wang2004image} are less consistent with visual quality and humans' evaluation than learning-based metrics such as FID~\cite{heusel2017gans}, LPIPS~\cite{zhang2018perceptual}, and FloLPIPS~\cite{danier2022flolpips}. There are examples shown in~\cite{lyu2024frame} indicating that PSNR/SSIM are inconsistent with visual qualities in VFI task, and we also include examples in Sec.~\ref{sec:results}. Thus, FID, LPIPS, and FloLPIPS are selected as the evaluation metrics. FID and LPIPS measure perceptual similarity by Fréchet distance or normalized distances between features extracted by deep learning models. FloLPIPS, developed on top of LPIPS, accounts for motion consistency. Recent state-of-the-art methods including PerVFI (2024)~\cite{wu2024perception}, Consec. BB (2024)~\cite{lyu2024frame}, LDMVFI (2024)~\cite{danier2024ldmvfi}, EMA-VFI (2023)~\cite{zhang2023extracting}, AMT (2023)~\cite{licvpr23amt}, UPR-Net (2023)~\cite{jin2023unified}, IFRNet (2022)~\cite{Kong_2022_CVPR}, VFIformer (2022)~\cite{lu2022video}, and MCVD (2022)~\cite{voleti2022mcvd} are selected as our baselines. For models with different versions on the number of parameters, the largest version is chosen. For completeness of our experiment, we include PSNR/SSIM and implementation details in our \textcolor{blue}{Supplementary Material Sec.~\ref{sec: PSNR}}.

\subsection{Experimental Results}
\label{sec:results}

\textbf{Quantitative Evaluation.} The quantitative results are shown in Tab.~\ref{tab:results}, with 10 sampling steps in diffusion. Notably, PerVFI uses a combination of flow estimators, RAFT~\cite {teed2020raft} and GMflow~\cite{xu2022gmflow}, to guide their model. These flow estimators are trained with three datasets: FlyingThings3D~\cite{MIFDB16}, FlyingChairs~\cite{DFIB15}, and Sintel~\cite{Butler:ECCV:2012}, where Sintel and Flyingthings3D contain high-resolution images. They additionally train their model on the Vimeo 90K Triplets dataset~\cite{xue2019video}. In contrast, all other methods are trained purely on the Vimeo 90K Triplets dataset without high-resolution images. Therefore, \textbf{the training scale of PerVFI is almost doubled with high-resolution data included} compared to other methods. To ensure a fair comparison, we display the results of PerVFI in \colorbox{gray!30}{gray background} and exclude them from ranking. However, we still include it as a reference to assess our method's capability. Even with a much smaller training scale than PerVFI, our method still achieves a better performance than PerVFI in most datasets and metrics. 

Under a fair comparison, our method achieves the best performance in all metrics and datasets, as shown in Tab.~\ref{tab:results}. \textbf{In the most challenging datasets}: SNU-FILM extreme~\cite{choi2020channel} and Xiph-4K~\cite{Niklaus_CVPR_2020}, our proposed methods achieve approximately \textbf{20}\% improvements in FID and FloLPIPS over the second-best results. In relatively easier datasets like Xiph-2K, our method achieves around \textbf{20}\% improvements in all metrics over the second-best result. In some datasets like DAVIS and Xiph-2K, we notice that newer methods can sometimes underperform older methods (like EMAVFI'23 vs IFRNet'22), but our method consistently outperforms others, indicating a \textbf{strong generalization capability} over different datasets. Moreover, we observe that the improvement margin increases when the task gets harder (such as SNU-FILM), indicating \textbf{our strong capability for challenging cases}, which is more crucial than incrementing quantitative scores on near-perfect results such as SNU-FILM-easy and medium.

\textbf{Runtime Analysis.} Other than interpolation quality, our method achieves good efficiency, shown in the last column of Tab.~\ref{tab:results}. Under the same number of sampling steps (10 in our setup), our method achieves \textbf{2.3}$\times$ speedup over Consec. BB and \textbf{4.3}$\times$ speedup over LDMVFI. Our method also achieves \textbf{2.2}$\times$ speedup than recent non-diffusion SOTA: PerVFI. This is an important step toward efficiency of diffusion-based methods, as previous diffusion-based methods are inefficient. Moreover, even PerVFI contains a larger training scale, our method achieves comparable results and much faster inference speed.

\begin{figure*}[ht]
\centering
\includegraphics[width=\textwidth]{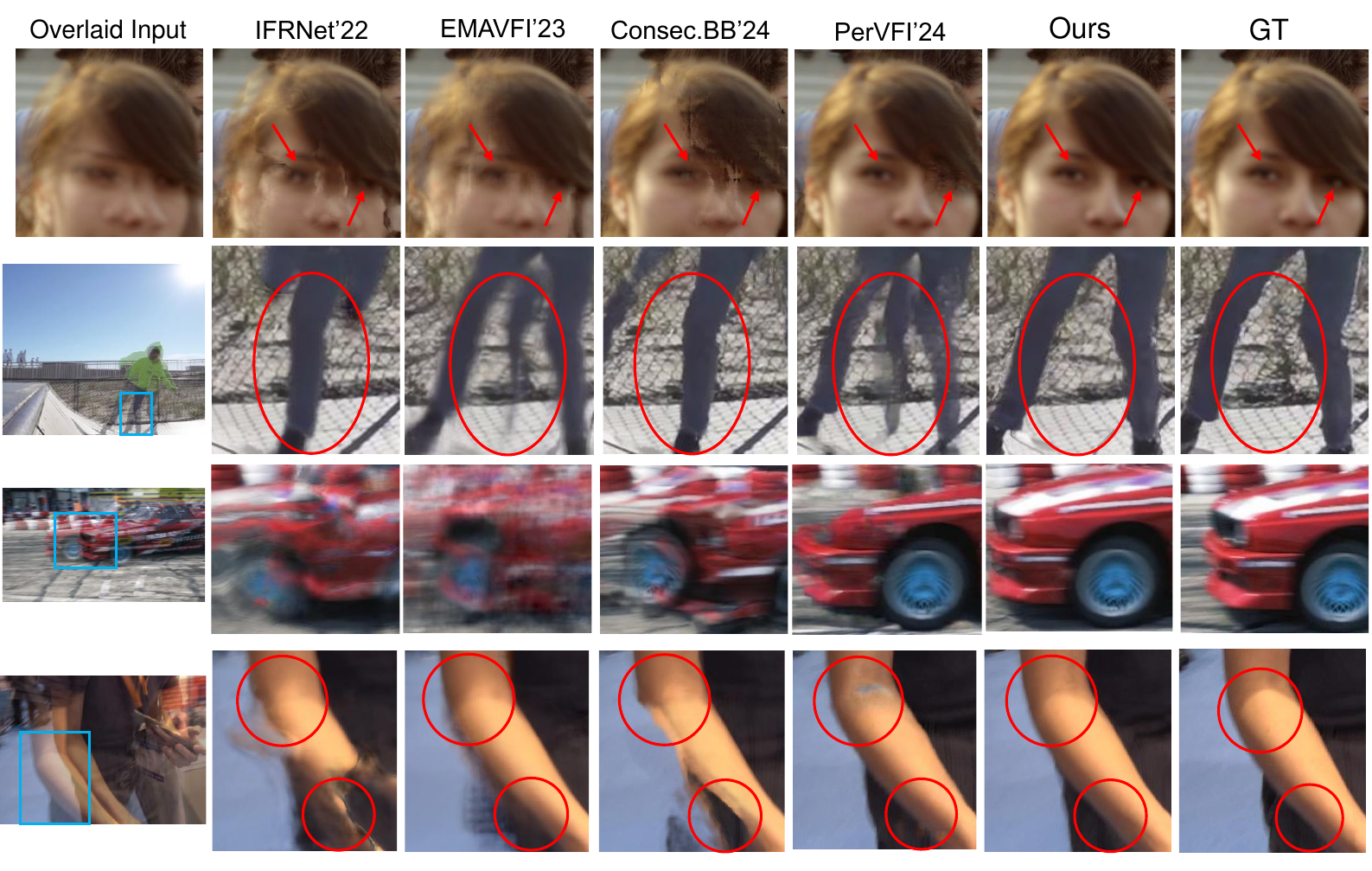}
\captionsetup{font=footnotesize,aboveskip=1pt}
\caption{\textbf{Qualitative Comparison between our method and recent SOTAs.} The leftmost image is the overlaid image of $I_0$ and $I_1$ (blended image). Areas with drastic motion changes are cropped with \textcolor{iccvblue} {blue} boxes to better visualize the results. \textcolor{red}{Red} circles and boxes indicate the area where we perform significantly better. Our method achieves better visual quality than recent SOTAs. Additional qualitative results are included in the \textcolor{blue}{Supplementary Material Sec.~\ref{sec: more qual}}.\vspace{-5mm}}

\label{fig:qual}
\end{figure*}

\textbf{Qualitative Evaluation.} We select Consec. BB~\cite{lyu2024frame}, PerVFI~\cite{wu2024perception}, EMAVFI~\cite{zhang2023extracting}, and IFRNet~\cite{Kong_2022_CVPR} to visually compare the interpolation results since they achieve strong results in challenging cases like SNU-FILM-extreme. The qualitative results are shown in Fig.~\ref{fig:qual}, with examples selected from challenging cases in SNU-FILM-extreme, Xiph-4K, and DAVIS. Overlaid Input means the blended image of $I_0$ and $I_1$. In the first row, results from IFRNet, EMAVFI, and Consec. BB are largely distorted, and the result in PerVFI contains a distorted eye. Our method aligns with the ground truth. In the second row, our method is the only one that does not contain ``the third leg" in the middle. The person's right leg in $I_0$ and left leg in $I_1$ are at almost the same position, resulting in an artificial ``third leg" generated by other methods. In the third row, IFRNet, EMAVFI, and Consec. BB generate distorted results, and the vehicle generated by PerVFI misses some parts. However, our method generates a high-quality image. In the fourth row, EMAVFI generates blurred results, and IFRNet and Consec. BB generate distorted results. PerVFI, though without large distortion, contains artifacts shown in the \textcolor{red}{red} circles, but our result is realistic. We additionally include $8\times$ interpolation results in our \textcolor{blue}{Supplementary Material Sec.~\ref{sec: more qual}}.

\begin{figure}[t]
\captionsetup{font=footnotesize,aboveskip=3pt}
\centering
\includegraphics[width=0.47\textwidth]{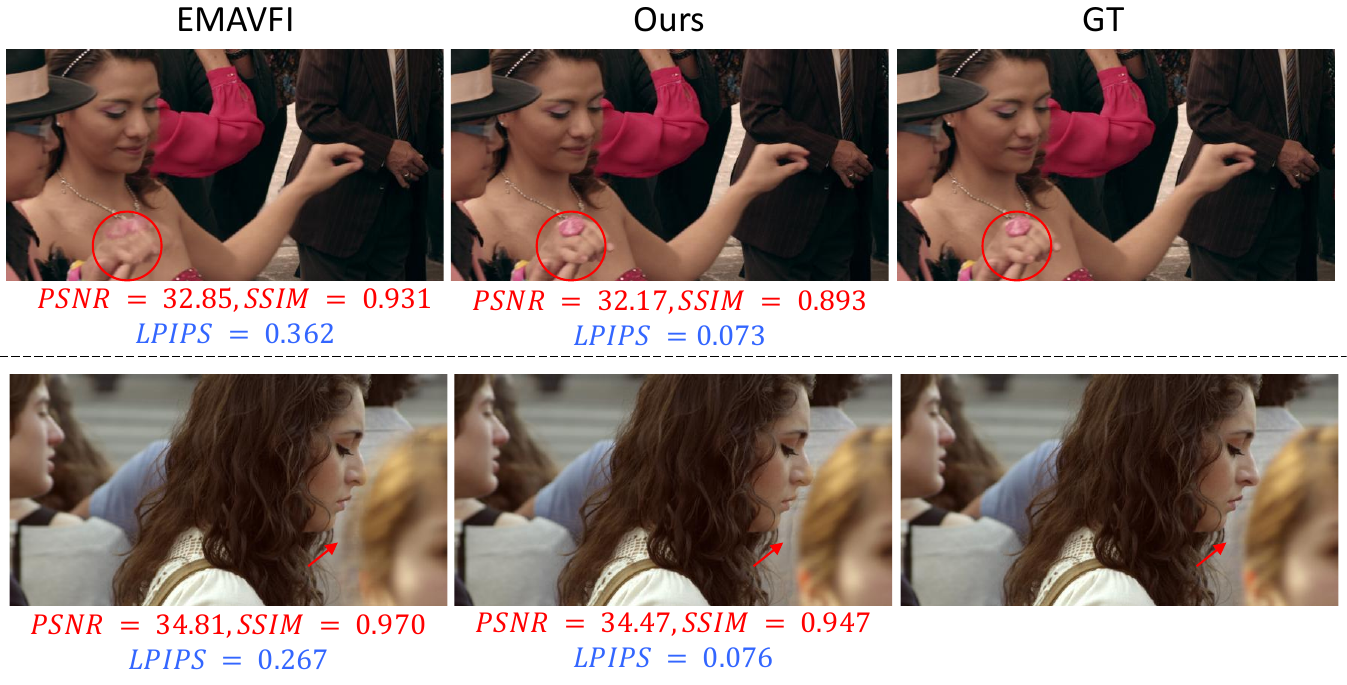}
\caption{\textbf{Inconsistency between PSNR/SSIM and Visual Quality.} \textcolor{red}{Red} circles and arrows indicate where the results from EMAVFI are distorted.\vspace{-5mm}}
\label{fig:inconsistent}
\end{figure}

\textbf{Inconsistency Between PSNR/SSIM and Visual Quality.} Examples in Fig.~\ref{fig:inconsistent} indicate that PSNR/SSIM are inconsistent with visual quality. In the first row, the female's hand in the \textcolor{red}{red} circle is blurred and distorted in the result of EMAVFI, but our result is consistent with ground truth. In the second row, there is distortion (pointed by \textcolor{red}{arrow}) on hairs in the result of EMAVFI, and ours is also consistent with the ground truth. However, in both examples, the PSNR/SSIM in EMAVFI is better, and our LPIPS is better, indicating that LPIPS is consistent with visual quality while PSNR/SSIM are not.

\textbf{Training Cost of Diffusion Based Models.} As claimed, our method achieves 20$\times$ fewer model parameters and requires 9000$\times$ less training data than video-based diffusion models~\cite{jain2024video,yang2024vibidsampler,shen2024dreammover}. We include quantitative results of the number of parameters and training data size of both image-diffusion-based methods and video-based diffusion methods in Tab.~\ref{tab:param}. Since MCVD~\cite{voleti2022mcvd} gets much worse performance than recent SOTAs (see Tab.~\ref{tab:results}), we exclude it for comparison. VIDIM~\cite{jain2024video} is trained with WebVid-10M~\cite{Bain21} and private videos, where WebVid-10M is an extension of WebVid-2M, which has 13K hours of videos, corresponding to 1.4T frames. Vimeo Triplet 90K contains 51K triplets, corresponding to 153K frames, and therefore, our method gets at least \textbf{9000}$\times$ less training data than VIDIM. The number of parameters, based on Tab.~\ref{tab:param}, is 20$\times$ fewer. Even comparing with image-based diffusion methods~\cite{danier2024ldmvfi,lyu2024frame}, our number of parameters is over 3$\times$ fewer.

\subsection{Ablation Studies}
\label{sec:ablatoin}

\begin{figure}[t]
\captionsetup{font=footnotesize,aboveskip=3pt}
\centering
\includegraphics[width=0.47\textwidth]{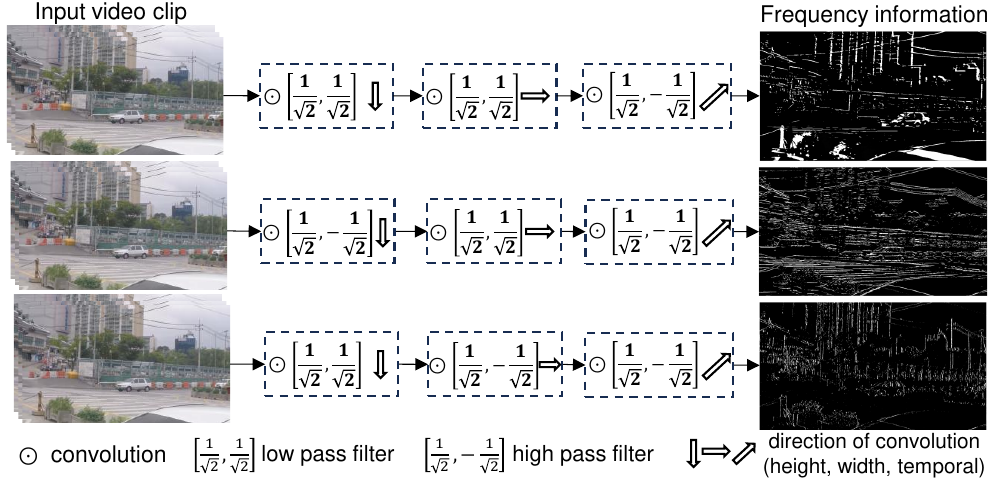}
\caption{\textbf{Visualization of 3D Wavelet Transform.}The 3D wavelet transform can extract frequency information along different directions (time, height, width). We visualize results by applying a high pass filter in the time dimension with a combination of (low, low), (low, high), and (high, low) filters in spatial dimensions.\vspace{-3mm}}
\label{fig:freq}
\end{figure}

\textbf{Temporal Aware Design.} We include ablation studies in Tab.~\ref{tab:ablation} to indicate the effectiveness of each component. We start with our method and gradually remove 3D wavelet, replace cross-attention aggregation by average pooling, remove temporal attention, and replace 3D convolution by 2D. We also include Ours$^\dagger$, where we replace Image Encoder by a temporal aware design in our full model. The performance gets worse since multi-level features are changed due to zero replacement in this setup, indicating the necessity of the Image Encoder. We additionally show three example visualizations of 3D wavelet transform in Fig.~\ref{fig:freq}. We apply high-pass and low-pass filters as convolution kernels to the input videos. By switching between high-pass and low-pass filters in temporal, height, and width dimensions, we can effectively extract 8 different frequency maps. Our temporal autoencoder plays the primary role, with the 3D wavelet feature gating contributing pixel-level information that further enhances the results.

\begin{table}[t]
    \centering
    \captionsetup{font=footnotesize,aboveskip=3pt}
    \caption{Training and inference cost of recent diffusion-based methods. \textcolor{red}{$^*$} means VIDIM~\cite{jain2024video} is trained additionally with private datasets.\textcolor{blue}{$\dagger$} means the performance of MCVD is much worse than SOTAs of recent years. Runtime is measured in seconds per frame with $480\times720$ resolution. N/A indicates that VIDIM~\cite{jain2024video} is not open-sourced, and NF indicates that Dreammover~\cite{shen2024dreammover} needs fine-tuning for every single inference example, which takes minutes.}
    \label{tab:param}
        \resizebox{.47\textwidth}{!}{
        \begin{tabular}{cccccc}
        Method&Input type&\# Parameters&\# Training Data & Pretrained & Runtime\\
        \midrule
        LDMVFI'24~\cite{danier2024ldmvfi}&Images&439.0M& 51K Triplets& \ding{55} & 2.48 \\
        Consec. BB'24~\cite{lyu2024frame}&Images&146.4M& 51K Triplets&\ding{55}&1.62\\
        \midrule
        MCVD'22~\cite{voleti2022mcvd}&Videos&27.3M\textcolor{blue}{$\dagger$}& 51K Triplets\textcolor{blue}{$\dagger$}&\ding{55}&52.55\\
        VIDIM'24~\cite{jain2024video}&Videos&$>$1B& $>$10M videos\textcolor{red}{$^*$}&\ding{55}&N/A\\
        Dreammover'24~\cite{shen2024dreammover}&Videos&943.2M& 244.7M Videos&\ding{51} & NF\\
        ViBiDSampler'24~\cite{yang2024vibidsampler}&Videos&943.2M& 244.7M Videos&\ding{51} & 8.48\\
        
        Ours&Videos&\textbf{46.7M}& \textbf{51K Triplets}&\ding{55}&0.69\\

        \bottomrule
    \end{tabular}
  }
  \vspace{-3mm}
\end{table}

\begin{table}[t]
    \centering
    \captionsetup{font=footnotesize,aboveskip=3pt}
    \caption{Ablation studies. The result is in FID. The - signs mean that the component is removed. The removal is one at a time. $^\dagger$ means that we use the full version of our method, but the Image Encoder is temporal-aware. $^\ddagger$ means we disable feature sharing from $^\dagger$.}
    \label{tab:ablation}
        \resizebox{.47\textwidth}{!}{
        \begin{tabular}{cccc}
        \toprule
        Dataset&Xiph-4K&Xiph-2K&SNU-FILM-extreme\\
        \midrule
        Ours&19.114&9.901&29.868\\
        - 3D Wavelet&19.247 & 10.092 & 30.717\\
        - Cross-attention aggregation&19.663&10.499&30.903\\
        - Temporal attention&19.944&10.911&32.061\\
        - 3D Convolution &23.481&12.679&33.155\\
        Ours$^\dagger$ &22.731&13.410&34.982\\
        Ours$^\ddagger$ &20.004&10.769&32.018\\
        \bottomrule
    \end{tabular}
  
  }
  \vspace{-3mm}
\end{table}

\begin{figure}[t]
\centering
\captionsetup{font=footnotesize,aboveskip=3pt}
\includegraphics[width=0.47\textwidth]{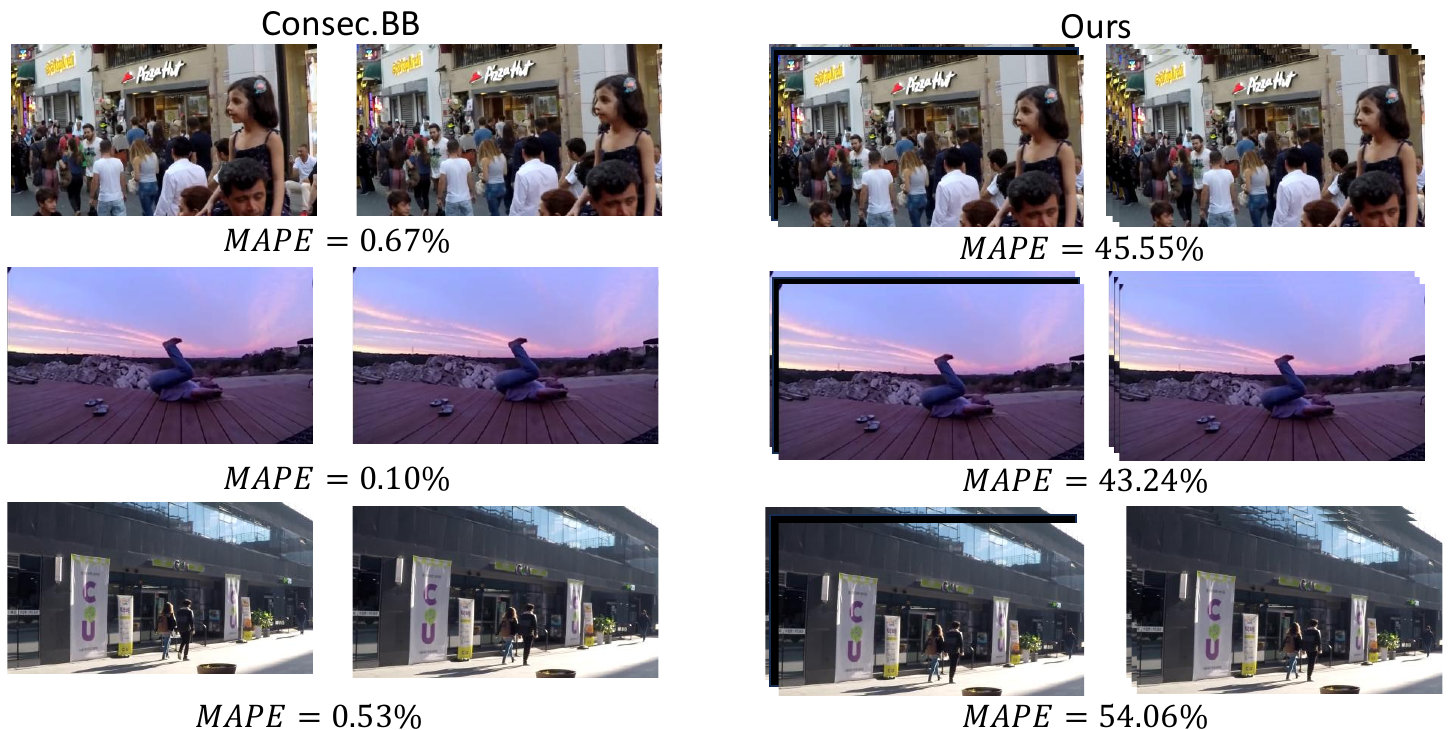}
\caption{$MAPE(\mathcal{E}(I_n),\mathcal{E}(I_0))$ in the setup of Consec. BB~\cite{lyu2024frame} and $MAPE(\mathcal{E}(V),\mathcal{E}(\tilde{V}))$ in our method. $\tilde{V}$ is the video clip with the intermediate frame replaced with 0s. The MAPE in Consec. BB is less than 1\%, resulting in a rough identity transformation. In our method, MAPE is 40-50\%, and therefore the Brownian Bridge learns to reduce this gap.\vspace{-5mm}}

\label{fig:mse}
\end{figure}

\textbf{Distribution Shift in Brownian Bridge.} The Brownian Bridge in Consec. BB~\cite{lyu2024frame} connects the encoder features of adjacent frames. Since adjacent frames are similar, features connected with Brownian Bridge Diffusion are almost identical, and therefore the diffusion is approximately an identity mapping. However, our design mitigates this issue by replacing $I_n$ in the video clip $V$ (results denoted as $\tilde{V}$). Following proposition~\ref{propostion}, we do a t-test for our method and Consec. BB~\cite{lyu2024frame}. The dataset is selected to be SNU-FILM extreme~\cite{choi2020channel} for the t-test. The t-statistic computed in our method is more than 21, which is much larger than the threshold of significance level 0.001: 3.291. This means that we can reject $H_0$. However, the t-statistic computed in the setup of Consec. BB is about 0.0001, where we cannot reject $H_0$. We show some examples in Fig.~\ref{fig:mse}. The Mean Absolute Percentage Error (MAPE) is used to measure the distribution shift of encoded features, where $MAPE(x,y) = \mathbb{E}|\frac{x-y}{x}|$, indicating the average percentage change. In these examples, we can see that the MAPE in Consec. BB is close to 0, but our method gives a large MAPE, experimentally showing the distributional shift.

\section{Conclusion}
In this paper, we introduce our TLB-VFI to extract temporal information in both latent space (temporal blocks) and pixel space (3D wavelet feature gating). With such a design, our method achieves state-of-the-art results and solves the limitations of recent diffusion-based methods. Our method is highly flexible as video diffusion-based models that we can handle more than 3 frames as input even though our method is trained with triplets. Meanwhile, we do not require large-scale training like video diffusion-based models because we take advantage of flow estimation to guide the generation. 

{
    \small
    \bibliographystyle{ieeenat_fullname}
    \bibliography{main}
}
\clearpage
\setcounter{page}{1}
\maketitlesupplementary

\begin{figure*}[t]
\centering
\includegraphics[width=\textwidth]{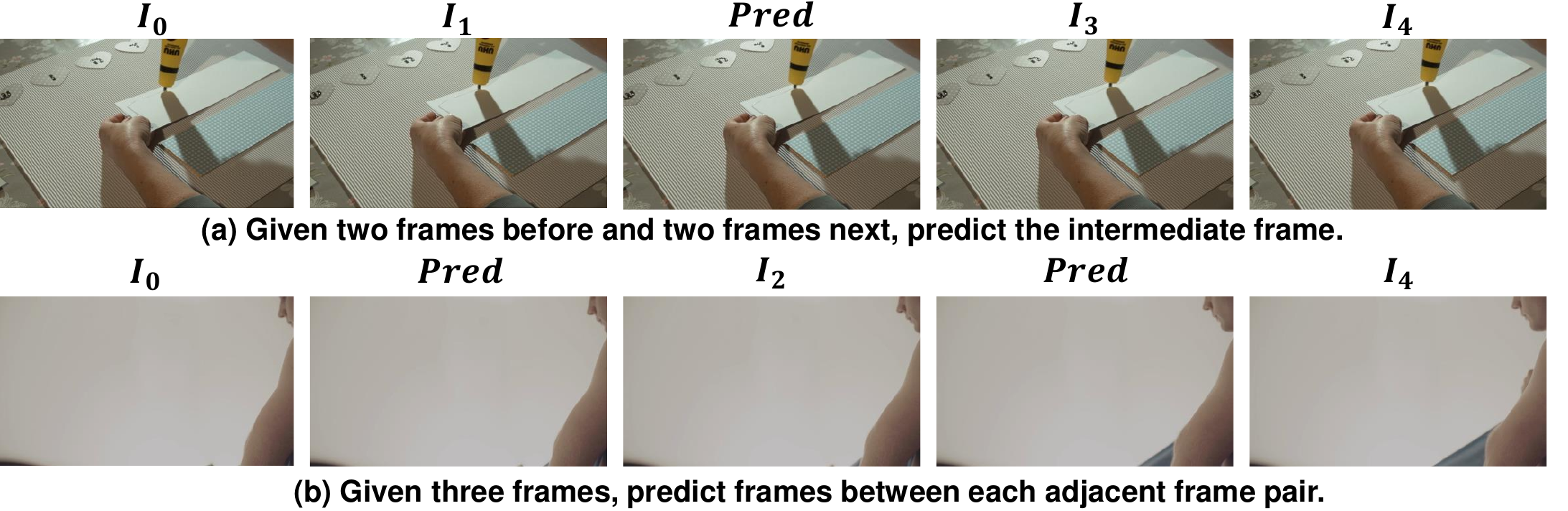}
\captionsetup{font=footnotesize,aboveskip=1pt}
\caption{(a) Given four neighboring frames $I_0,I_1,I_3,I_4$, we can predict the intermediate frame $I_2$. (b) Given a sequence of frames $I_0,I_2,I_4$, we can predict the intermediate frame between each adjacent pair $I_1,I_3$.}

\label{fig:mult}
\end{figure*}

\section{Overview}
The supplementary material is structured as follows:

\begin{itemize}
    \item Sec.~\ref{sec: flow} includes optical basics: what are optical flows and warping.
    \item Sec.~\ref{sec: PSNR} contains PSNR/SSIM evaluated on our selected datasets.
    \item Sec.~\ref{sec: more qual} contains additional qualitative results.
    \item Sec.~\ref{sec:implement} contains implementation details.
    \item Sec.~\ref{sec:wavelet} contains additional details on 3D wavelet transforms.
    \item Sec.~\ref{sec:proof} contains the proof to our proposition in Sec. 3.3 of our main paper. 
\end{itemize}

\section{Optical Flow Basics}
\label{sec: flow}
Optical flow is the pixel-wise movement from frame to frame. If we have two images $I_0$ and $I_1$, and for a given pixel $I_0[i,j]$ the corresponding pixel appears in $I_1$ $I_1[i',j']$ then $flow(I_0,I_1)[i,j]$ is $[i'-i,j'-j]$, indicating the pixel movement. Warping is to move each pixel according to the movements defined by optical flows. Importantly, optical flows do not explicitly estimate the motion speed, and motion speed is implicitly contained in training data. If all training data consists of constant speed motion, and the test data contains motion speed that is not evenly distributed between two frames, the predicted position will not align. This would be an interesting research problem but out of scope of our research. An example is the third row of Fig. 3 in our main paper, where our method and PerVFI predict basically the same location but the ground truth location is different. One possible explanation is that the vehicle is accelerating, but the location that our method and PerVFI predicts is based on a constant speed. As a result, at the first half of the time interval between $I_0,I_1$, the vehicle is slow so the ground truth is closer to the first frame, while our method and PerVFI make it approximately right in the middle.

\section{Additional Results}

\begin{table*}[t]
    \centering
    \caption{Quantitative results (PSNR/SSIM) on test datasets (the higher the better).OOM indicates that the inference with one image exceeds the 24GB GPU memory of an Nvidia RTX A5000 GPU.}
    \label{tab:results PSNR}
    \resizebox{\textwidth}{!}{
        \begin{tabular}{cccccccc}
        \toprule
        \multirow{2}{*}{Methods}&\multirow{2}{*}{Xiph-4K} & \multirow{2}{*}{Xiph-2K}& \multirow{2}{*}{DAVIS}& \multicolumn{4}{c}{SNU-FILM} \\
        \cmidrule{5-8}
        & & & &easy & medium & hard&extreme\\
        &PSNR/SSIM&PSNR/SSIM&PSNR/SSIM&PSNR/SSIM&PSNR/SSIM&PSNR/SSIM&PSNR/SSIM\\
        \midrule
        
        MCVD'22~\cite{voleti2022mcvd} &OOM &OOM& 18.946/0.705 &  22.201/0.828 & 21.488/0.812 &  20.314/0.766&18.464/0.694\\

        VFIformer'22~\cite{lu2022video} &  OOM & OOM & 26.241/0.850 &40.130/0.991&36.090/0.980  & 30.670/0.938 & 25.430/0.864 \\
        
        IFRNet'22~\cite{Kong_2022_CVPR} & 33.970/0.943&36.570/0.966&27.313/0.877 &40.100/0.991&36.120/0.980 &30.630/0.937 &25.270/0.861 \\

        AMT'23~\cite{licvpr23amt} & 34.653/0.949 &36.415/0.967 & 27.234/0.877&39.880/0.991  &36.120/0.981&30.780/0.939 &25.430/0.865 \\
        
        UPR-Net'23~\cite{jin2023unified} &33.647/0.946&36.749/0.967& 26.894/0.870 & 40.440/0.991& 36.290/0.980&30.860/0.938& 25.630/0.864\\
        
        EMA-VFI'23~\cite{zhang2023extracting} &34.698/0.948 &36.935/0.967& 27.111/0.871&39.980/0.991&  36.090/0.980& 30.940/0.939&25.690/0.866 \\
        
        LDMVFI'24~\cite{danier2024ldmvfi} & OOM&OOM&   25.073/0.819&38.890 0.988&33.975/0.971&29.144/0.911&23.349 0.827\\

        PerVFI'24~\cite{wu2024perception} &32.395/0.926&34.741/0.953&26.502/0.866&38.065/0.986&34.588/0.973&29.821/0.928&25.033/0.854\\

        Consec.BB~\cite{lyu2024frame} &32.153/0.927&34.964/0.956 &26.391/0.858 &39.637/0.990& 34.886/0.974& 29.615/0.929 &24.376/0.848 \\

        \midrule
        Ours &32.441/0.928&35.748/0.959 &26.272/0.860 &39.460/0.990& 35.308/0.977& 29.529/0.929 &24.513/0.847  \\

        \bottomrule
    \end{tabular}
  }
\end{table*}

\subsection{Results in PSNR/SSIM}
\label{sec: PSNR}
We include the results in PSNR/SSIM on our selected datasets in Tab.~\ref{tab:results PSNR}. We can see that PSNR/SSIM tends to be unstable and not correlated to visual qualities for methods in 2024. For example, our method underperforms Consec. BB in Xiph~\cite{Niklaus_CVPR_2020} dataset but our visual comparisons and LPIPS/FloLPIPS/FID indicate that our method is better. Similarly, PerVFI underperforms Consec. BB in Xiph-2K, but its LPIPS/FLoLPIPS/FID is much better.

\subsection{Additional Qualitative Results}
\label{sec: more qual}

\begin{figure}[t]
\centering
\includegraphics[width=0.47\textwidth]{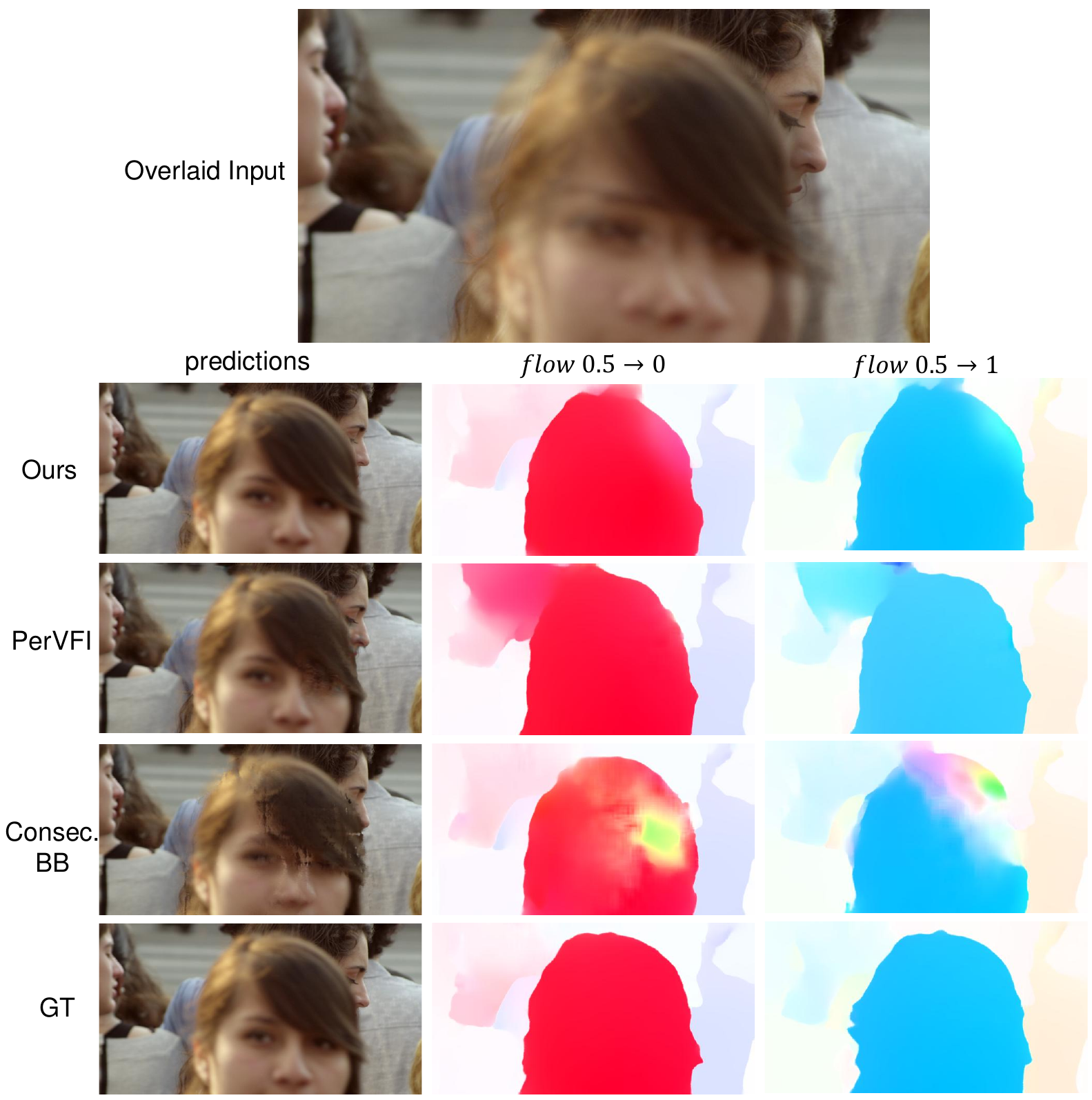}
\captionsetup{font=footnotesize,aboveskip=1pt}
\caption{Visualization of optical flows.}

\label{fig:flow}
\end{figure}

\begin{figure*}[t]
\centering
\includegraphics[width=\textwidth]{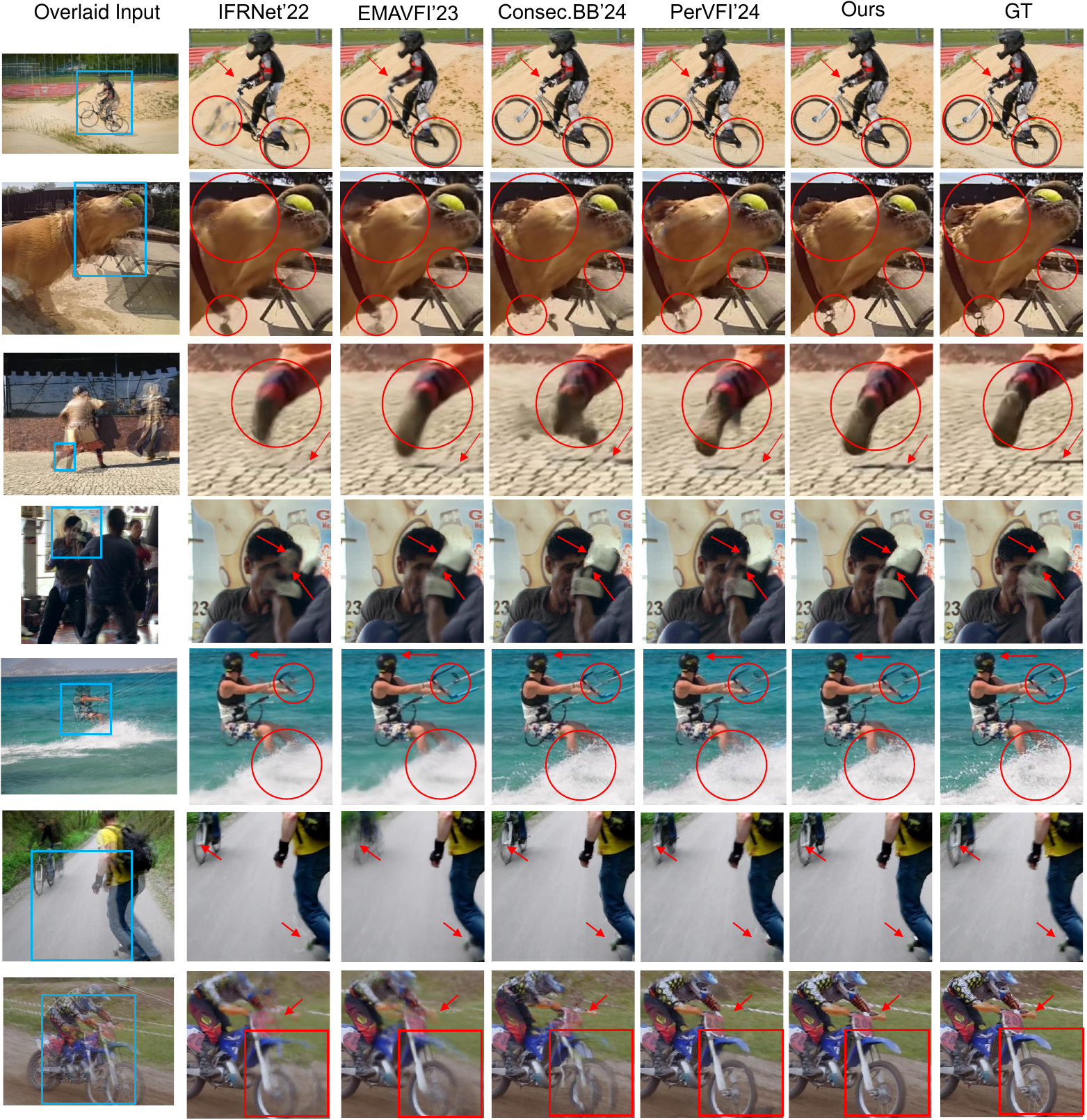}
\captionsetup{font=footnotesize,aboveskip=1pt}
\caption{\textbf{Additional qualitative comparison between our method and recent SOTAs.} The leftmost image is the overlaid image of $I_0$ and $I_1$ (blended image). Images inside \textcolor{iccvblue} {blue} boxes contain drastic motion changes and are cropped out to show details of interpolation results. \textcolor{red}{Red} circles, boxes, and arrows indicate the area where we significantly perform better. Our method achieves better visual quality than recent SOTAs.\vspace{-5mm}}

\label{fig:qual2}
\end{figure*}

\begin{figure*}[t]
\centering
\includegraphics[width=\textwidth]{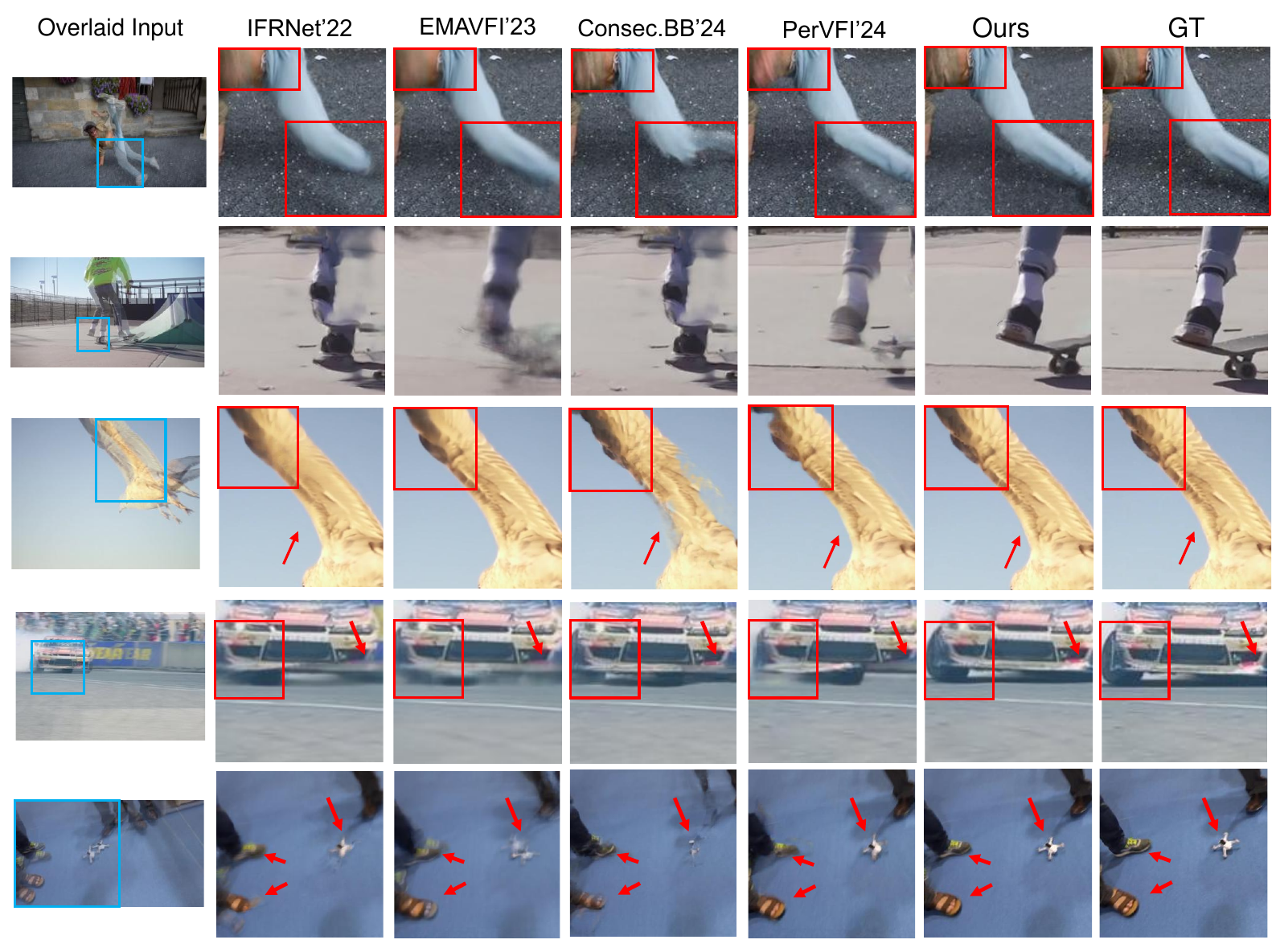}
\captionsetup{font=footnotesize,aboveskip=1pt}
\caption{\textbf{Additional qualitative comparison between our method and recent SOTAs.} The leftmost image is the overlaid image of $I_0$ and $I_1$ (blended image). Images inside \textcolor{iccvblue} {blue} boxes contain drastic motion changes and are cropped out to show details of interpolation results. \textcolor{red}{Red} circles, boxes, and arrows indicate the area where we significantly perform better. Our method achieves better visual quality than recent SOTAs.\vspace{-5mm}}

\label{fig:qual3}
\end{figure*}

\begin{figure*}[t]
\centering
\includegraphics[width=\textwidth]{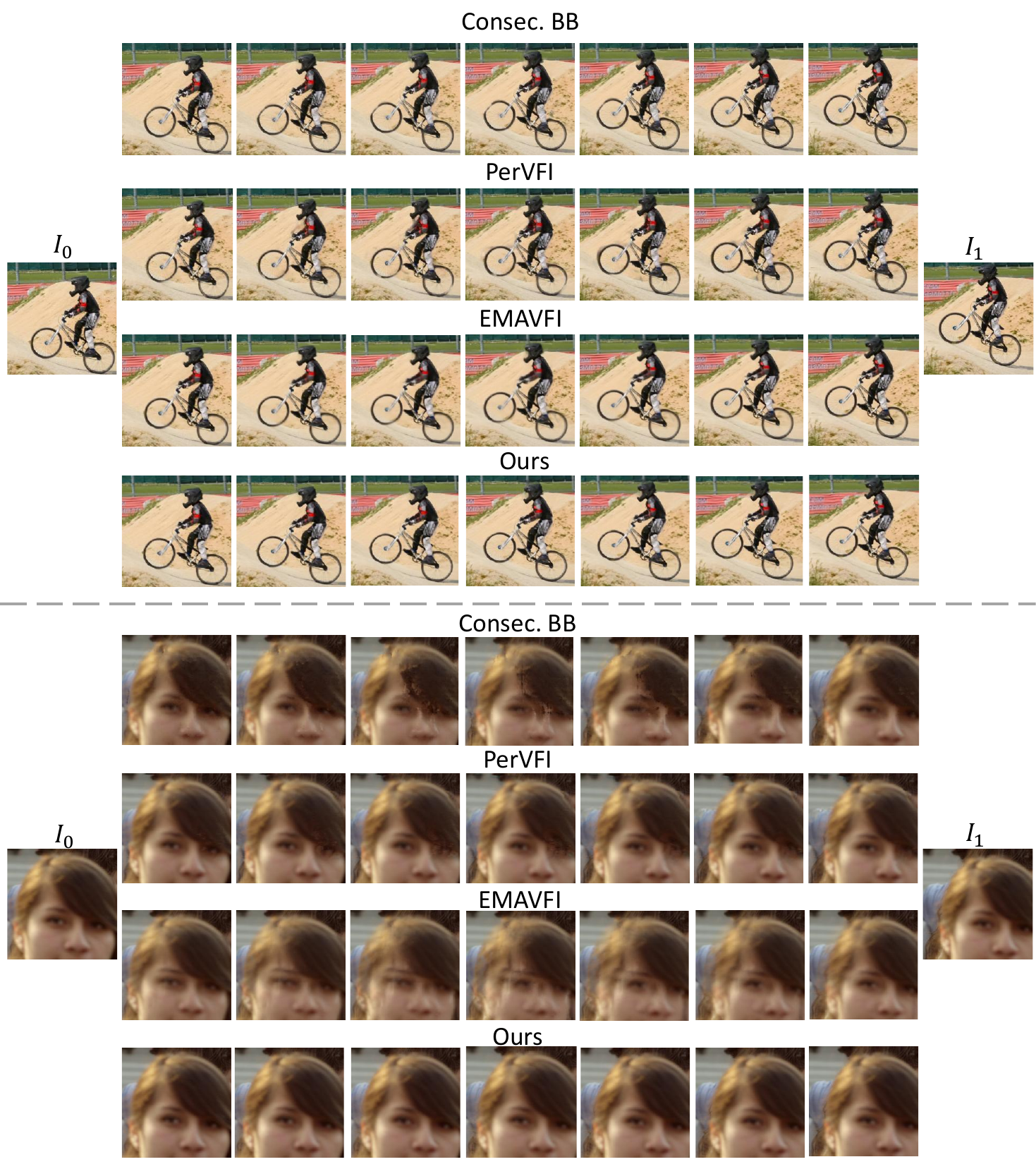}
\captionsetup{font=footnotesize,aboveskip=1pt}
\caption{\textbf{Visual comparison of 8x$\times$ interpolation results.} We include a visual comparison of 8$\times$ interpolation between our method and PerVFI. \textcolor{red}{Red} arrows indicate where our method is visually better. Additional comparisons (in video form) are provided in our \textcolor{blue}{\href{https://zonglinl.github.io/tlbvfi_page}{Project Page}}.\vspace{-5mm}}

\label{fig:interp8}
\end{figure*}

\textbf{More Input Frames.} Our method is highly flexible and can take more than three input frames in one forward call. An example is shown in Fig.~\ref{fig:mult}, where our model can receives five frames $I_0,I_1,I_2,I_3,I_4$ and treat either $I_2$ or $I_1,I_3$ as target. This enables us to do only one run of the sampling process for the second scenario, where LDMVFI~\cite{danier2024ldmvfi} and Consec. BB~\cite{lyu2024frame} needs to sample twice. To achieve this, we only need to replace the second and fourth frames by zeros and send the video clip to the autoencoder, and the diffusion model only needs one sampling process to obtain latents for the decoder. However, LDMVFI and Consec. BB needs to sample frame 2 and 4 separately.

\textbf{Additional Visual Comparisons.} We include additional visual comparisons in Fig.~\ref{fig:qual2} and Fig.~\ref{fig:qual3}, where examples are selected from SNU-FILM extreme~\cite{choi2020channel}, DAVIS~\cite{Perazzi_CVPR_2016}, and Xiph-4K~\cite{Niklaus_CVPR_2020}. Our method achieves the best visual quality. Other methods exhibit distortion, blurring, or artifacts in their generation, but our method does not. \textcolor{red}{Red} circles and squares emphasize the area where our method achieves better quality. We encourage reviewers to do $500\%$ zoom-in to see the results as many results contain multiple details in one frame.

\textbf{8$\times$ Interpolation.} 8$\times$ interpolation is interpolating 7 frames between $I_0,I_1$, which can be done iteratively. When motion change is large, $2\times$ interpolation does not provide a good video clip and therefore we need to interpolate more frames. We include two examples of $8\times$ interpolation results in Fig.~\ref{fig:interp8} just for reference. It is better to visualize $8\times$ with videos, and therefore we include more examples compared with more methods in our \textcolor{blue}{\href{https://zonglinl.github.io/tlbvfi_page}{Project Page}}. 8$\times$ interpolation is to interpolate 7 intermediate frames between $I_0$ and $I_1$ (i.e. only first and last frames are provided), which can be done iteratively. The upper example is taken from DAVIS~\cite{Perazzi_CVPR_2016} and the latter one from Xiph-4K~\cite{Niklaus_CVPR_2020}. The \textcolor{blue}{\href{https://zonglinl.github.io/tlbvfi_page}{Project Page}} contains results from SNU-FILM extreme~\cite{choi2020channel}, DAVIS~\cite{Perazzi_CVPR_2016}, and Xiph-4K~\cite{Niklaus_CVPR_2020}. In the upper examples of Fig.~\ref{fig:interp8}, bicycle tires interpolated by PerVFI~\cite{wu2024perception} are missing in some frames. In the lower example, the woman's right eye becomes an artifact in PerVFI.

\textbf{Flow visualization.} We visualize the optical flow from interpolated results ($\hat{I}_n$) to neighboring frames $I_0,I_1$, shown in Fig.~\ref{fig:flow}. Our primary contribution is the temporal aware latent Brownian Bridge diffusion framework instead of advancements in optical flows in other works~\cite{wu2024perception,zhang2023extracting,licvpr23amt}, so we directly adopt the flow estimation architecture from Consec. BB~\cite{lyu2024frame}. Though the flow estimator is the same architecture as Consec. BB, our temporal design can implicitly improve it through back-propagation (see first and third row).

\section{Additional Details}

\label{add_detail}

\subsection{Implementation Details}
\label{sec:implement}
\textbf{Flow Estimator.} Optical flow estimation is not our research purpose, so we use the same architecture of flow estimator in Consec. BB~\cite{lyu2024frame} and trained together with our autoencoder. The code for differentiable warping is available at~\cite{lu2022video,lyu2024frame}.

\textbf{Autoencoder.} The autoencoder is based on the VQ version of LDM~\cite{rombach2022high}. It consists of 5 levels of image encoder and decoder, resulting in a 32$\times$ downsampling rate. Image decoders contain output channels of 64,128,128,128,256, respectively (reverse for decoder). Between the image encoder and decoder, there are four 3D convolutions with spatiotemporal attention (the last one is cross-attention)~\cite{vaswani2017attention}, where a VQ-Layer is inserted after the second 3D conv + attention. The channel dimension is 256. The VQ-Layer quantizes features into 3 channels. To predict masks $M$ and residual $\Delta$, we use sigmoid activation to normalize the output. The autoencoder is trained with Adam optimizer~\cite{Adam} and a learning rate of $10^{-5}$ for 35 epochs. The training loss is L1 loss and LPIPS loss, following LDM.

\textbf{Brownian Bridge Diffusion.} The Brownian Bridge Diffusion is implemented with 3D denoising U-Net~\cite{rombach2022high} with channel dimension 32 and 3 downsample blocks as well as 3 upsample blocks, where the optimizer is Adam~\cite{Adam} with a learning rate of $10^{-4}$ and the model is trained for 50 epochs. The $T$ for the diffusion process is set to 2. The training loss is MSE loss.

\begin{algorithm}
\caption{Diffusion Training Algorithm}
\label{alg:train}
\begin{algorithmic}[1]
\State Let $\mathcal{E}$ be the encoder part of our autoencoder.
\For{$i = 1$ to $N_{training\text{ }steps}$}
\State Sample $t\sim ContinuousUniform(0,T)$.
\State Sample $[I_0,I_n,I_1]$ from Dataset. 
\State $x_0 = \mathcal{E}([I_0,I_n,I_1])$.
\State Compute $x_t$ with Eq.~\eqref{eq:BBDM forward}.
\State Take gradient step on $MSE(\epsilon_{\theta}(x_t),x_t-x_0)$
\EndFor
\end{algorithmic}
\end{algorithm}

\begin{algorithm}
\caption{Diffusion Sampling Algorithm}
\label{alg:sample}
\begin{algorithmic}[1]
\State Let $\mathcal{E}$ be the encoder part of our autoencoder.
\State Initialize $t=T,x_t=\mathcal{E}([I_0,0,I_1])$ 
\While{$t>0$}
\State Predict $x_t - x_0$ with $\epsilon_\theta(x_t)$
\State Sample $x_s$ with Eq.~\eqref{BBDM_sample}
\State $t \gets s$, $x_t\gets x_s$
\EndWhile
\end{algorithmic}
\end{algorithm}

The training algorithm is shown in Algorithm~\ref{alg:train}, and the sampling algorithm is shown in Algorithm~\ref{alg:sample}.

\subsection{3D Wavelet Details}
\label{sec:wavelet}
The 3D wavelet transform can be considered as a convolution layer with two types of filter: high-pass filter $[\frac{1}{\sqrt{2}},-\frac{1}{\sqrt{2}}]$ and low-pass filter $[{\frac{1}{\sqrt{2}}},-\frac{1}{\sqrt{2}}]$.The input videos (with 3 frames) are converted to grayscale, and filters are applied in height, width, and temporal dimensions respectively. There are 8 different combinations of filters: $[HHH,HHL,HLH,HLL,LHH,LHL,LLH,LLL]$, corresponding to height, width, and temporal dimensions. We use ``same padding" along height and width to keep the feature map size unchanged and use ``no padding" along the temporal dimension, both with a stride of 1. Therefore, after the first extraction, it provides 8 different feature maps, where each feature map contains 2 temporal channels (the convolution reduces the temporal dimension by 1). The $LLL$ is further extracted, resulting in 8 feature maps with only 1 temporal dimension. Feature maps other than $LLL$ are concatenated in the temporal dimension, and we consider the temporal dimension as ``channels" for model input to extract latent features. Therefore, we have a feature map with 21 channels as input.

\subsection{Proof}
\label{sec:proof}
We include the proof of our proposition in Sec. 3.3 of the main paper here:

\begin{proof}
    If $H_0$ is rejected, then it is statistically significant to conclude that $\mathbf{x}_T - \mathbf{x}_0 \neq 0$. Therefore, we can use a simple induction to prove this. Recall that we have the diffusion and sampling processes defined as:

    \begin{equation}
    \small
    \label{eq:BBDM forward supp}
    q(\mathbf{x}_t|\mathbf{x}_0,\mathbf{x}_T) = \mathcal{N}\left(\frac{t}{T}\mathbf{x}_0 + (1-\frac{t}{T})\mathbf{x}_T,\frac{t(T-t)}{T}\mathbf{I}\right).
    \end{equation}
    
    \begin{equation}
    \label{eq: BBDM_sample spp}
    p_\theta(\mathbf{x}_{s}|\mathbf{x}_t,\mathbf{x}_T) = \mathcal{N}\left(\mathbf{x}_t - \frac{\Delta_t
    }{t}(\mathbf{x}_t - \mathbf{x}_0),\frac{s\Delta_t}{t}\mathbf{I}\right).
    \end{equation}

    \begin{enumerate}
        \item Based on Eq.~\eqref{eq: BBDM_sample spp} in the main paper, suppose that at a given time step $t$, $\mathbf{x}_t - \mathbf{x}_0 \neq 0$, then the expectation of sampled latent at any previous step $s$ is $\mathbb{E}(\mathbf{x}_s|\mathbf{x}_t)=\frac{s}{t}(\mathbf{x}_t - \mathbf{x}_0) + \mathbf{x}_0$. 
        \item By the inductive assumption, $\mathbb{E}(\mathbf{x}_s|\mathbf{x}_t)= \mathbf{x}_0 + \delta,$ where $\delta \neq 0 \iff \mathbb{E}(\mathbf{x}_s|\mathbf{x}_t) \neq \mathbf{x}_0$.
        \item Then, on expectation, we conclude that $\mathbf{x}_s|\mathbf{x}_t \neq \mathbf{x}_0$. Note that this is especially important in DDIM~\cite{song2021denoising} sampling because the variance term is removed, in which case we can directly conclude that $\mathbf{x}_s|\mathbf{x}_t \neq \mathbf{x}_0$ without expectation.
        \item Therefore, we can prove this proposition by the above induction because the sampling process is discretized into finite steps.
    \end{enumerate}
    As a result, the sampling process is not an identity map.
\end{proof}

On the other hand, the sampling process is trivial because it does not change the expectation, which can be achieved with an identity map.

\clearpage


\end{document}